\documentclass[11pt,a4paper]{article}
\usepackage[hyperref]{emnlp2019}
\usepackage{times}
\usepackage{latexsym}
\usepackage{algorithm}
\usepackage[noend]{algpseudocode}
\usepackage{url}
\usepackage{stackengine}
\setstackEOL{\\}
\usepackage{amsmath}
\usepackage{amssymb}
\usepackage{subcaption}
\usepackage{multirow}
\usepackage{multicol}
\usepackage{color}
\usepackage{graphicx}
\usepackage{booktabs}

\aclfinalcopy 

\newcommand{\hs}[1]{\textcolor{blue}{Haitian: #1}} 
\newcommand{\pat}[1]{\textcolor{red}{Pat: #1}} 
\newcommand{\gray}[1]{\textcolor{gray}{}} 

\newcommand\blfootnote[1]{%
  \begingroup
  \renewcommand\thefootnote{}\footnote{#1}%
  \addtocounter{footnote}{-1}%
  \endgroup
}

\title{Facts as Experts: Adaptable and Interpretable Neural Memory \\ over Symbolic Knowledge}

\author{Pat Verga*, Haitian Sun*, Livio Baldini Soares, William W. Cohen \\
    Google Research \\
   \texttt{\{patverga, haitiansun, liviobs, wcohen\}@google.com} \\
  }

\date{}

\begin{document}
\maketitle
\blfootnote{*Equal contribution}
\begin{abstract}
Massive language models are the core of modern NLP modeling and have been shown to encode impressive amounts of commonsense and factual information.  However, that knowledge exists only within the latent parameters of the model, inaccessible to inspection and interpretation, and even worse, factual information memorized from the training corpora is likely to become stale as the world changes.  Knowledge stored as parameters will also inevitably exhibit all of the biases inherent in the source materials. To address these problems, we develop a neural language model that includes an explicit interface between symbolically interpretable factual information and subsymbolic neural knowledge.  We show that this model dramatically improves performance on two knowledge-intensive question-answering tasks.  More interestingly, the model can be updated without re-training by manipulating its symbolic representations. In particular this model allows us to add new facts and overwrite existing ones in ways that are not possible for earlier models.
\end{abstract}

\section{Introduction \label{sec:intro}}
Over the past several years, large pretrained language models (LMs) \citep{peters2018deep, devlin2019bert, raffel2019exploring} have shifted the NLP modeling paradigm from approaches based on pipelines of task-specific architectures to those based on pretraining followed by fine-tuning, where a large language model discovers useful linguistic properties of syntax and semantics through massive self-supervised training, and then small amounts of task specific training data are used to fine-tune that model (perhaps with small architectural modifications). 
More recently, similar approaches have been explored for knowledge representation and reasoning (KRR) with researchers asking questions like `Language Models as Knowledge Bases?' \citep{petroni2019language}. Results suggest that \citep{roberts2020much, brown2020language} the answer is a resounding `sort of' \citep{poerner2019bert}: while language models can be coerced to answer factual queries, they still lack many of the properties that knowledge bases typically have.
In particular, when evaluating LM-as-KRR models there are three explanations for why a model outputs a correct answer; 1) The model has successfully performed some reasoning or generalization required to make a novel inference, 2) the dataset contains some statistical biases that the model is exploiting, or 3) the model has memorized the exact answer, potentially from pre-training data that overlaps with the test cases.\footnote{This is a real possibility: for example, the T5 training data contains a large portion of the sources from which TriviaQA was derived, and attempts at avoiding leakage in GPT3 by looking at large ngram exact match do not account for trivial surface form changes.}.  In short, knowledge encoded only in a LM's parameters is generally opaque.

To address these problems, we propose an interface between explicit, symbolically bound memories and sub-symbolic distributed neural models. In addition to making more of a language model's behavior interpretable, our approach has several other important benefits. First, there is a massive amount of useful information that has been created and curated in structured databases. Sometimes this information either does not occur in text at all (such as a new product that hasn't come out yet) or is very difficult to interpret from the text (such as in scientific, technical, or legal documents). In our framework, new knowledge can be inserted by updating the symbolically bound memory.  
Second, pre-trained language models appear to require training on very large corpora to obtain good factual coverage---and the massive web corpora required by these data-hungry models contain huge amounts of sexist, racist, and incorrect assertions \citep{bolukbasi2016man,sun2019mitigating}. Our approach makes it possible to obtain better factual coverage of assertions chosen from selected trusted sources, by inserting this trusted factual content into the symbolic memory.

We propose to incorporate an external fact memory into a neural language model. This model forms its predictions by integrating contextual embeddings with retrieved knowledge from an external memory, where those memories are bound to symbolic facts which can be added and modified. We evaluate our model's performance empirically on two benchmark question answering datasets; FreebaseQA and WebQuestionsSP (section \ref{sec:results_full}). In section \ref{sec:results_inject}, we show how we can inject new memories at inference time enabling our model to correctly answer questions about pairs of entities that were never observed in the pretraining text corpus. Finally, in section \ref{sec:results_update} we examine to what extent our model is capable of iteratively updating by overwriting prior memories with new facts. We modify facts such that they actually contradict the original pretraining data, and show that our model is capable of answering correspondingly modified question answer pairs. In these experiments we show that end users can inject new knowledge and change existing facts by manipulating only the symbolically bound memories without retraining any parameters of the model.


\section{Related Work \label{sec:related_work}}
Knowledge bases (KBs) have been a core component of AI since the beginning of the field \citep{newell1956logic, newell1959}. Widely available public KBs have been invaluable in research and industry \citep{bollacker2008freebase, auer2007dbpedia} and many companies have created massive KBs as the backbones of their most important products \citep{googlekg2012, dong_2017}. 

While traditional KBs were purely symbolic, recent advances in large language models trained through self supervision \citep{peters2018deep, devlin2019bert, raffel2019exploring, brown2020language} have been shown to encode an impressive amount of factual information. This has led to research on the extent to which a neural language model can serve as a KB \citep{roberts2020much, petroni2019language}, and other research on how to best evaluate the factual knowledge in language models \citep{poerner2019bert}.

While large LMs appear to absorb KB-like information as a preproduct of pretraining, there has also been many prior approaches proposed that explicitly embed symbolic knowledge representations into neural embedding space. Various neural-symbolic methods have attempted to unify these two extremes \citep{pinkas1991symmetric,de2011neural,besold2017neural} including many cognitive architectures which used hybrid symbolic and subsymbolic systems \citep{laird2017standard}, and more recently, compositional query languages for embedding KBs that are similar to symbolic KB query languages \citep{cohen2017tensorlog,hamilton2018embedding,ren2020query2box,cohen2020scalable}. One system especially related to our proposal is EmQL \citep{sun2020guessing}, which includes a construct quite similar to the ``fact memory'' used in our Facts-As-Experts model.  Unlike this work, however, EmQL did not embed its fact memory into a language model, which can be finetuned for many NLP tasks: instead EmQL must be used with task-specific query templates and integrated into some task-specific architecture.

More recently, the past decade has seen huge amount of work on knowledge base embeddings \citep{bordes2013translating, lin2015learning, trouillon2017knowledge, dettmers2018convolutional} which enable generalization through similarities between learned embeddings. This  idea has also been extended with works looking at ways of incorporating raw text and symbolic KGs into a shared embedding space \citep{riedel2013relation, verga2016multilingual}, to be jointly reasoned over \citep{sun2018open, sun2019pullnet}, or to treat text as a replacement for a knowledge base \citep{dhingra2019differentiable}. 

Large external memories have been incorporated into different types of memory networks operating over latent parameters \citep{weston2014memory, miller2016key}, entity memories \citep{henaff2016tracking, fevry2020entities}, relations \citep{logan2019barack}, and embedded text passages \citep{guu2020realm, lewis2020retrieval}. Our work directly extends one of these models, the Entities-as-Experts (EaE) model \citep{fevry2020entities}, one of several models that inject knowledge of entities by constructing a memory containing embedded entity representations.  Unlike prior models, EaE learns entity representations end-to-end, rather than using representations from a separately-trained KB embedding system \citep{logan2019barack}.
Our work extends EaE by introducing a symbolic memory of triples which is constructed from these learned entity representations, and as in EaE, the entity representations are learned end-to-end.

\section{Model \label{sec:model}}


\subsection{Facts-as-Experts (FaE)}
\begin{figure*}[h!]
  \includegraphics[width=\textwidth]{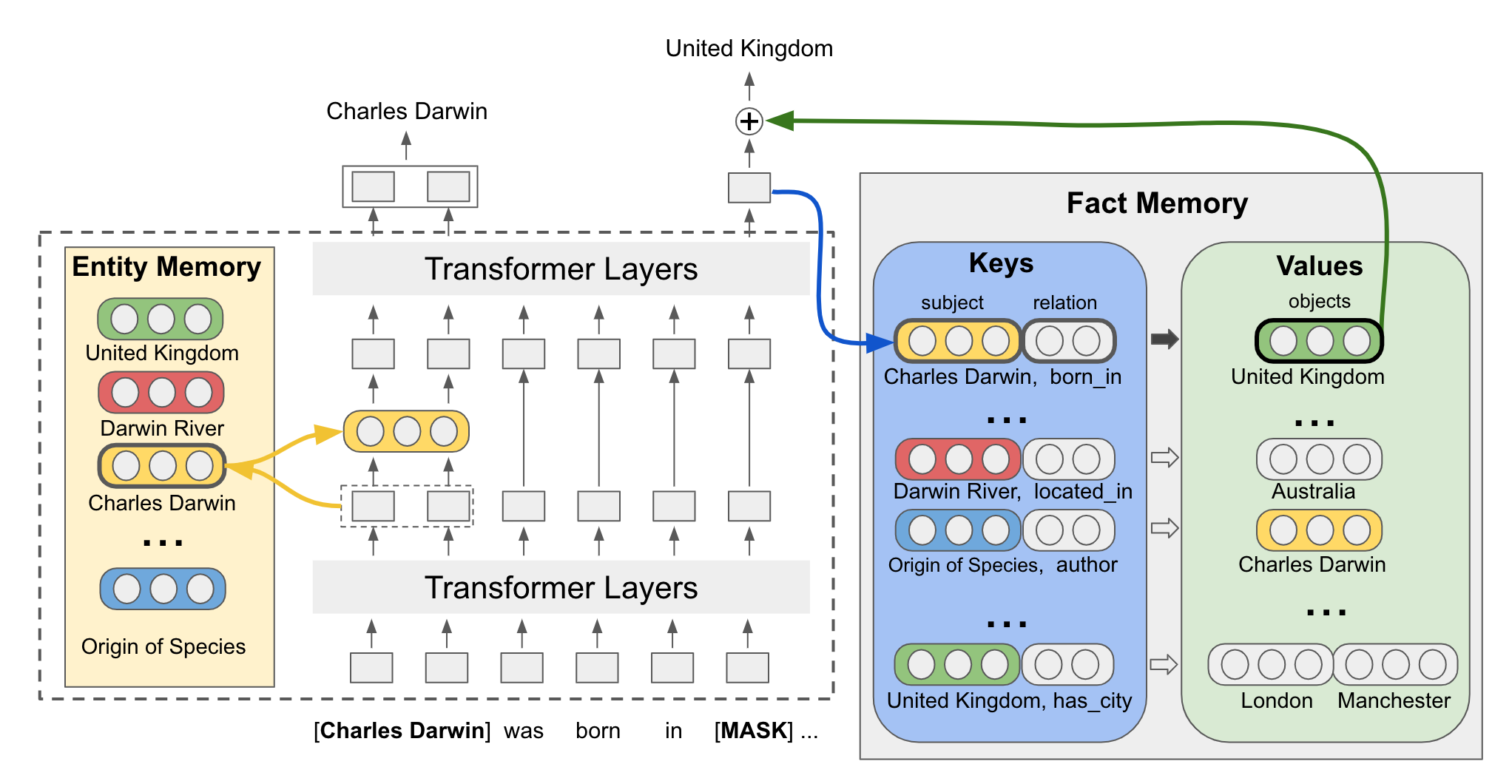}
  \caption{\textbf{Facts-as-Experts model architecture.} The model takes a piece of text (a question during fine-tuning or arbitrary text during pre-training) and first contextually encodes it with an entity enriched transformer. The part of the model within the dashed line is exactly the Entities-as-Experts model from \citet{fevry2020entities}. The model uses the contextually encoded MASK token as a query to the fact memory. In this case, the contextual query chooses the fact key (\textit{Charles Darwin, born\_in}) which returns the a set of values \{\textit{United Kingdom}\} (The value set can be multiple entity objects such as the case from calling the key [\textit{United Kingdom, has\_city}]) . The returned object representation is incorporated back into the context in order to make the final prediction. Note that the entities in facts (both in keys and values) are shared with the EaE entity memory. 
  \label{fig:fae_model}}
\end{figure*}

Our Facts-as-Experts (FaE) model (see Figure \ref{fig:fae_model}) builds an interface between a neural language model and a symbolic knowledge graph.  This model builds on the recently-proposed Entities as Experts (EaE) language model \citet{fevry2020entities}, which extends the same transformer \citep{vaswani2017attention} architecture of BERT \citep{devlin2019bert} with an additional external memory for entities.  After training EaE, the embedding associated with an entity will (ideally) capture information about the textual context in which that entity appears, and by inference, the entity's semantic properties.  In FaE, we include an additional memory called a \emph{fact memory}, which encodes triples from a symbolic KB. Each triple is constructed compositionally from the EaE-learned embeddings of the entities that comprise it.  This fact memory is represented with a key-value memory, and can be used to retrieve entities given their properties in the KB.  This combination results in a neural language model which learns to access information in a the symbolic knowledge graph.





\subsection{Definitions \label{sec:model_definition}}
We represent a Knowledge Base $\mathcal{K}$ as a set of triples $(s, r, o)$ where $s, o \in \mathcal{E}$ are the subject and object entities and $r \in \mathcal{R}$ is the relation, where $\mathcal{E}$ and $\mathcal{R}$ are pre-defined vocabularies of entities and relations in the knowledge base $\mathcal{K}$. A text corpus $\mathcal{C}$ is a collection of paragraphs\footnote{We use the term paragraph to describe a text span that is roughly paragraph length (128 token pieces in our experiments). In reality the text spans do not follow paragraph boundaries.} $\{p_1, \dots, p_{|C|}\}$. Let $\mathcal{M}$ be the set of entity mentions in the corpus $\mathcal{C}$. A mention $m_i$ is defined as $(e_m, s^p_m, t^p_m)$, i.e. entity $e_m$ is mentioned in paragraph $p$ starting from the token at position $s^p_m$ and ends on $t^p_m$. Since we don't consider multi-paragraph operations in this paper, we will usually drop the superscript $p$ and use $s_m$ and $t_m$ for brevity.

\subsection{Input \label{sec:model_input}}
The input to our model is a piece of text which can be either a question in the case of fine tuning (see section \ref{sec:model_finetune}) or an arbitrary span as in pre-training (see section \ref{sec:model_pretrain}). Our pretraining input is constructed as cloze-type Question Answering (QA) task. Formally, given a paragraph $p = \{w_1, \dots, w_{|p|}\}$ with mentions $\{m_1, \dots, m_n\}$, we pick a mention $m_i$ and replace all tokens from $s_{m_i}$ to $t_{m_i}$ with a special $[\textnormal{MASK}]$ token. We consider the entity in $\mathcal{E}$ named by the masked entity to be the answer to the cloze question $q$. Mentions in the paragraph other than this masked entity are referred as below as \textit{context mentions}. For example, in the cloze question,
\{\textit{`Charles', `Darwin', `was', `born', `in', [MASK], [MASK], `in', `1809', `.', `His', `proposition', $\dots$}\}, ``Charles Darwin'' is a context entity in mention $m_1$ = (\textnormal{`Charles Darwin'}, 1, 2), and ``United Kingdom'' is the answer entity in the masked mention $m_{\textnormal{ans}}$ = (\textnormal{`United Kingdom'}, 6, 7). 


Our model learns to jointly link entities from context mentions $m_i$ using entity-aware contextual embeddings (\S \ref{sec:model_eae}) and predict answer entities
using knowledge-enhanced embeddings (\S \ref{sec:model_fae}).  This process  will be introduced in more detail in the following sections.


\subsection{Entity-aware Contextual Embeddings} \label{sec:model_eae}
We follow the Entities-as-Experts (EaE) \citep{fevry2020entities} model to train an external \textit{entity memory}. The EaE model is illustrated in the left part of Figure \ref{fig:fae_model}. This model interleaves standard Transformer layers with layers that access an entity memory (see \citet{vaswani2017attention} for details on the transformer architecture). EaE inputs a paragraph (or question) containing unlinked entities with known boundaries\footnote{\citet{fevry2020entities} also showed the model is capable of learning to predict these boundaries. For simplicity, in this work we assume they are given.} (i.e., the index of the start and end of each mention is provided, but the identity of the entity mentioned is not.)  Given a question $q = \{w_1, \dots, w_{|q|}\}$ with a list of context mentions $m_i$ = ($e_{m_i}$, $s_{m_i}$, $t_{m_i}$) and the answer $e_{\textnormal{ans}}$ from the masked mention $m_{\textnormal{ans}}$ = ($e_{\textnormal{ans}}$, $s_{\textnormal{ans}}$, $t_{\textnormal{ans}}$), the contextual embedding $\textbf{h}_i^{(l)}$ is the output at the $i$'th token of the $l$'th intermediate transformer layer.  
\begin{align*}
    \textbf{h}_i^{(l)}, \dots, \textbf{h}_{|q|}^{(l)} &= \textnormal{Transformer}(\{w_1, \dots, w_{|q|}\})
\end{align*}
These contextual embeddings are used to compute query vectors that interface with an external entity memory $\textbf{E} \in \mathbb{R}^{|\mathcal{E}| \times d_e}$, which is a large matrix containing a vector for each entity in $\mathcal{E}$.
To construct a query vector, we concatenate the context embeddings for the mention $m_i$'s start and end tokens, $\textbf{h}^{(l)}_{s_{m_i}}$ and $\textbf{h}^{(l)}_{t_{m_i}}$, and project them into the entity's embedding space. We compute the attention weights over the embeddings of the full entity vocabulary, and use this to produce the attention-weighted sum of entity embeddings $\textbf{u}^{l}_{m_i}$.  This result is then projected back to the dimension of the contextual token embeddings, and added to what would have been the input to the next layer of the Transformer. 
\begin{align}
    \textbf{h}^{(l)}_{m_i} &= \textbf{W}_e^T [\textbf{h}^{(l)}_{s_{m_i}}; \textbf{h}^{(l)}_{t_{m_i}}] \label{eq:entity_access_vector} \\
    \textbf{u}^{(l)}_{m_i} &= \textnormal{softmax} (\textbf{h}^{(l)}_{m_i}, \textbf{E}) \times \textbf{E} \\
    \Tilde{\textbf{h}}^{(l+1)}_j &= \textbf{h}^{(l)}_{j} + \textbf{W}_2^T \textbf{u}^{(l)}_{m_i},~~ s_{m_i} < j < t_{m_i}
\end{align}

Let $\textbf{h}^{(T)}_j$ be the contextual embedding of the $j$'th token after the final transformer layer $T$. Similar to the query construction in the intermediate transformer layer in Eq. \ref{eq:entity_access_vector}, EaE constructs the query vector $\textbf{h}^{(T)}_{m_i}$ for mention $m_i$ and use it to predict the context entities $\hat{e_{m_i}}$. This query vector is called an \textit{entity-aware contextual query} in the rest of this paper and denoted as $\textbf{c}_{m_i}$ for brevity. This query vector is trained with a cross-entopy loss against $\mathbb{I}_{e_{m_i}}$, the one-hot label of entity $e_{m_i}$.
\begin{align*}
    \hat{e}_{m_i} &= \textnormal{argmax}_{e_i \in \mathcal{E}} (\textbf{c}^T_{m_i} \textbf{e}_i) \\
    \textnormal{loss}_{\textnormal{ctx}} &= \textnormal{cross\_entropy} (\textnormal{softmax} (\textbf{c}_{m_i}, \textbf{E}), \mathbb{I}_{e_{m_i}})
\end{align*}

As shown in \citet{fevry2020entities}, supervision on the intermediate entity access is beneficial for learning entity-aware contextual embeddings. We compute an entity memory access loss using the intermediate query vector in Eq. \ref{eq:entity_access_vector}.
\begin{align*}
    \textnormal{loss}_{\textnormal{ent}} &= \textnormal{cross\_entropy} (\textnormal{softmax} (\textbf{h}^{(l)}_{m_i}, \textbf{E}), \mathbb{I}_{e_{m_i}})
\end{align*}


In pretraining the FaE model, we used a slightly different pre-training process than was used in EaE. In EaE, mentions in the same paragraphs are independently masked with some probability and jointly trained in one example.\footnote{EaE is also jointly trained on mention detection. Please refer to \citet{fevry2020entities} for more information.} In FaE, in addition to the randomly masked context mentions, FaE picks exactly one of the mentions and masks it. Predicting this masked entity requires additional access to the fact memory which will be discussed in the next section.


\subsection{Fact Memory \label{sec:model_fact_memory}}

FaE inherits the external entity memory $\mathbf{E}$ from the EaE model and adds another \textit{fact memory} which contains triples from the knowledge base $\mathcal{K}$  (see the right side of Figure \ref{fig:fae_model}). The fact memory shares its on entity representations with the entity memory embeddings in $\textbf{E}$, but each element of the fact memory corresponds to a symbolic substructure, namely a key-value pair $((s, r), \{o_1, \dots, o_n\})$.
The key $(s, r)$ is a (subject entity, relation) pair, and the corresponding value $\{o_1, \dots, o_n\}$ is the list of object entities associated with $s$ and $r$,  i.e. $(s, r, o_i) \in \mathcal{K}$ for $i=\{1, \dots, n\}$. Hence, conceptually, KB triples with the same subject entity and relation are grouped into a single element. We call the subject and relation pair $a_j = (s, r) \in A$ a \textit{head pair} and the list of objects $b_j = \{o_1, \dots, o_n\} \in B$ a \textit{tail set}\footnote{The size of the tail set $b_j$ can be large for a popular head pair $(s, r)$. In such case, we randomly select a few tails and drop the rest of them. The maximum size of the tail set is 32 in the experiments in this paper.}, and will encode $\mathcal{K}$  as a new structure  $\mathcal{K}' = (A, B)$, with $|A|=|B|$. Notice that $\mathcal{K}'$ contains same information as $\mathcal{K}$, but can be encoded as as a key-value memory: elements are scored using the keys $(s, r)$ from head pairs $A$, and values from the tail sets $B$ are returned. 

In more detail, we encode a head pair $a_j = (s, r) \in A$ in the embedding space as follows.
Let $\mathbf{E} \in \mathbb{R}^{|\mathcal{E}| \times d_e}$ be the entity embeddings trained in Sec \ref{sec:model_eae}, and $\mathbf{R} \in \mathbb{R}^{|\mathcal{R}| \times d_r}$ be embeddings of relations $\mathcal{R}$ in the knowledge base $\mathcal{K}$. We encode a head pair $a$ as:
$$
\textbf{a}_j = \textbf{W}_a^T ~[\textbf{s} ; \textbf{r}] ~\in \mathbb{R}^{d_a}
$$
\noindent where $\textbf{s} \in \mathbf{E}$ and $\textbf{r} \in \mathbf{R}$ are the embeddings of subject $s$ and relation $r$, and $\textbf{W}_a$ is a learned linear transformation matrix. $\mathbf{A} \in \mathbb{R}^{|A| \times d_a}$ is the embedding matrix of all head pairs in $A$. 

The query vector to access the fact memory is derived from contextual embeddings and projected to the same embedding space as the head pairs $\mathbf{A}$. For a masked mention $m_{\textnormal{ans}} = (e_{\textnormal{ans}}, s_{\textnormal{ans}}, t_{\textnormal{ans}})$, define a query vector
\begin{align} \label{eq:fact_query_vector}
    \textbf{v}_{m_{\textnormal{ans}}} &= \textbf{W}_f^T ~[\textbf{h}^{(T)}_{s_{\textnormal{ans}}}; \textbf{h}^{(T)}_{t_{\textnormal{ans}}}]
\end{align}
\noindent where $\textbf{h}^{(T)}_{s_{\textnormal{ans}}}$ and $\textbf{h}^{(T)}_{t_{\textnormal{ans}}}$ are the contextual embeddings at the start and end tokens for the mention $m_{\textnormal{ans}}$, and $\textbf{W}_f$ is the linear transformation matrix into the embedding space of head pairs $\mathbf{A}$.

Head pairs in $A$ are scored by the query vector $\textbf{v}_{m_{\textnormal{ans}}}$ and the top $k$ head pairs with the largest inner product scores are retrieved. This retrieval process on the fact memory is distantly supervised. We define a head pair to be a \textit{distantly supervised positive example} $a_{\textnormal{ds}} = (s, r)$ for a passage if its subject entity $s$ is named by a context mention $m_i$ and the masked entity $e_{\textnormal{ans}}$ is an element of the corresponding tail set, i.e. $e_{\textnormal{ans}} \in b_{\textnormal{ds}}$. In cases that no distantly supervised positive example exists for a passage, we introduce add a special example that retrieves a  ``null'' fact from the knowledge base, where the ``null'' fact has a special $s_\textnormal{null}$ head entity and special $r_\textnormal{null}$ relatio: i.e. $a_\textnormal{ds} = (s_\textnormal{null}, r_\textnormal{null})$ and its tail set is empty. This distant supervision is encoded by a loss function
\begin{align*}
    \textnormal{TOP}_k(\textbf{v}_{m_{\textnormal{ans}}}, \textbf{A}) = \textnormal{argmax}_{k, j \in \{1, \dots, |A|\}} \textbf{a}_j^T \textbf{v}_{m_{\textnormal{ans}}} \\
    \textnormal{loss}_{\textnormal{fact}} = \textnormal{cross\_entropy}(
        \textnormal{softmax}(\textbf{v}_{m_{\textnormal{ans}}}, \textbf{A}), \mathbb{I}_{a_{\textnormal{ds}}})
\end{align*}
\noindent Here the tail sets associated with the top $k$ scored head pairs, i.e. $\{b_j | j \in \textnormal{TOP}_k(\textbf{v}, \textbf{A})\}$, will be returned from the fact memory. We will discuss integrating the retrieved tail sets $b_j$'s to the context in the following section.

\subsection{Integrating Knowledge and Context \label{sec:model_fae}}
Tail sets retrieved from the fact memory are next aggregated and integrated with the contextual embeddings. Recall that a tail set $b_j$ returned from the fact memory is the set of entities $\{o_1, \dots, o_n\}$ s.t. $(s, r, o_i) \in \mathcal{K}$ for $i \in \{1, \dots, n\}$ with the associated $a_j = (s, r)$. Let $\textbf{o}_i \in \mathbf{E}$ be the embedding of entity $o_i$. We encode the returned tail set $\textbf{b}_j$ as a weighted centroid of the embeddings of entities in the tail set $b_j$.
$$
\textbf{b}_j = \sum_{o_i \in b_j} \alpha_i \textbf{o}_i ~\in \mathbb{R}^{d_e}
$$
\noindent where $\alpha_i$ is a context-dependent weight of the object entity $o_i$.  To compute the weights $\alpha_i$, we use a process similar to Eq. \ref{eq:fact_query_vector}, and we compute a second query vector $\textbf{z}_{m_{\textnormal{ans}}}$ to score the entities inside thee tail set $b_j$. 
The weights $\alpha_i$ are the softmax of the inner products between the query vector $\textbf{z}_{m_{\textnormal{ans}}}$ and the embeddings of entities in the tail set $b_j$.
\begin{align} 
    \textbf{z}_{m_{\textnormal{ans}}} &= \textbf{W}_b^T [\textbf{h}^{(T)}_{s_{\textnormal{ans}}}; \textbf{h}^{(T)}_{t_{\textnormal{ans}}}] \label{eq:tail_embs}\\
    \alpha_i &= \frac{\exp{(\textbf{o}_i^T ~ \textbf{z}_{m_{\textnormal{ans}}})}}{\sum_{o_l \in b_j} \exp{(\textbf{o}_l^T ~ \textbf{z}_{m_{\textnormal{ans}}})}}
\end{align}
\noindent where $\textbf{W}_b$ is yet another transformation matrix different from $\textbf{W}_e$ in Eq. \ref{eq:entity_access_vector} and $\textbf{W}_f$ in Eq. \ref{eq:fact_query_vector}. 
The top $k$ tail sets $\textbf{b}_j$ are further aggregated using weights $\beta_j$, which are the softmax of the retrieval (inner product) scores of the top $k$ head pairs $a_j$. This leads to a single vector $\textbf{f}_{m_{\textnormal{ans}}}$ that we call the \textit{knowledge embedding} for the masked mention $m_{\textnormal{ans}}$.
\begin{align} 
    \textbf{f}_{m_{\textnormal{ans}}} &= \sum_{j \in \textnormal{TOP}_k (\textbf{v}_{m_{\textnormal{ans}}}, \textbf{A})} \beta_j \textbf{b}_j \label{eq:knowledge-query}\\
    \beta_j &= \frac{\exp{(\textbf{a}_j^T ~ \textbf{v}_{m_{\textnormal{ans}}})}}{\sum_{t \in \textnormal{TOP}_k (\textbf{v}_{m_{\textnormal{ans}}}, \textbf{A})} \exp{(\textbf{a}_t^T ~ \textbf{v}_{m_{\textnormal{ans}}})}}
\end{align}



Intuitively $\textbf{f}_{m_{\textnormal{ans}}}$ is the result of retrieving a set of entities from the fact memory.  We expect FaE should learn to jointly use the contextual query  $\textbf{c}_{m_{\textnormal{ans}}}$ and knowledge query $\textbf{f}_{m_{\textnormal{ans}}}$ to predict the masked entity, i.e. use external knowledge retrieved from the fact memory if there exists an oracle head pair $a_{\textnormal{orc}} = (s, r)$ s.t. $e_{\textnormal{ans}} \in b_{\textnormal{orc}}$, or fall back to contextual query to make predictions otherwise. We compute the integrated query $\textbf{q}_{m_{\textnormal{ans}}}$ with a mixing weight $\lambda$. $\lambda$ is the probability of predicting the ``null'' head $a_{\textnormal{null}}$ in the fact memory access step, i.e. whether an oracle head pair $a_{\textnormal{orc}}$ exists in the knowledge base.
\begin{align*}
    \lambda &= P(y = a_{\textnormal{null}}) \\
    \textbf{q}_{m_{\textnormal{ans}}} &= \lambda \cdot \textbf{c}_{m_{\textnormal{ans}}} + (1 - \lambda) \cdot \textbf{f}_{m_{\textnormal{ans}}}
\end{align*}

The query vector $\textbf{q}_{m_{\textnormal{ans}}}$ is called a \textit{knowledge-enhanced contextual query}. This query vector finally is used to predict the masked entity. Again, we optimized it with a cross-entropy loss.
\begin{align*}
    \hat{e}_{\textnormal{ans}} &= \textnormal{argmax}_{e_i \in \mathcal{E}} (\textbf{q}^T_{m_{\textnormal{ans}}} \textbf{e}_i) \\
    \textnormal{loss}_{\textnormal{ans}} &= \textnormal{cross\_entropy} (\textnormal{softmax} (\textbf{q}_{m_{\textnormal{ans}}}, \textbf{E}), \mathbb{I}_{e_{\textnormal{ans}}})
\end{align*}

\subsection{Pretraining\label{sec:model_pretrain}}
FaE is jointly trained to predict context entities and the masked entity. Context entities are predicted using the contextual embeddings described in $\S$ \ref{sec:model_eae}; intermediate supervision with oracle entity linking labels is provided in the entity memory access step for context entities; the masked entity is predicted using the knowledge-enhanced contextual embeddings ($\S$ \ref{sec:model_fae}); and distant supervised fact labels are also provided at training time. The final training loss is the unweighted sum of the four losses:
$$
\textnormal{loss}_{\textnormal{pretrain}} = \textnormal{loss}_{\textnormal{ent}} + \textnormal{loss}_{\textnormal{ctx}} + \textnormal{loss}_{\textnormal{fact}} + \textnormal{loss}_{\textnormal{ans}}
$$

\subsection{Finetuning on Question Answering \label{sec:model_finetune}}
In the Open-domain Question Answering task, questions are posed in natural language, e.g. ``Where was Charles Darwin born?'', and answered by a sequence of tokens, e.g. ``United Kingdom''. In this paper, we focus on a subset of open-domain questions that are answerable using entities from a knowledge base. In the example above, the answer ``United Kingdom'' is an entity in  Wikidata whose identity is Q145. 

We convert an open-domain question to an input of FaE by appending the special [MASK] token to the end of the question, e.g. \{`Where', `was', `Charles', `Darwin', `born', `?', [MASK]\}. The task is to predict the entity named by mask. Here, ``Charles Darwin'' is a context entity, which is also referred to as \textit{question entity} in the finetuning QA task.

At finetuning time, entity embeddings $\mathbf{E}$ and relation embeddings $\mathbf{R}$ are fixed, and we finetune all transformer layers and the four transformation matrices: $\textbf{W}_a$, $\textbf{W}_b$, $\textbf{W}_e$, $\textbf{W}_f$. Parameters are tuned to optimize unweighted sum of the the fact memory retrieval loss $\textnormal{loss}_{\textnormal{fact}}$ and the final answer prediction loss $\textnormal{loss}_{\textnormal{ans}}$. If multiple answers are available, the training label $\mathbb{I}_{e_{\textnormal{ans}}}$ becomes a $k$-hot vector uniformly normalized across the answers.
$$
\textnormal{loss}_{\textnormal{finetune}} = \textnormal{loss}_{\textnormal{fact}} + \textnormal{loss}_{\textnormal{ans}}
$$

\section{Experiments \label{sec:results}}




\subsection{Datasets}
\begin{table}[h!]
    \small
    \centering
    \begin{tabular}{ll|ccc}
\toprule
        &                           & Full              &  Wikidata \\ 
        &                             & Dataset          &  Answerable   \\ 
\midrule
        & Train                       &  20358                      &  12535          \\
FreebaseQA    & Dev                         &  2308                      &  2464               \\
        & Test                        &  3996                      &  2440         \\
\midrule
        & Train &  2798                      &  1388      \\
WebQuestionsSP  & Dev &  300                  &  153     \\
        & Test  &  1639                 & 841      \\
\bottomrule
    \end{tabular}
    \caption{\textbf{Dataset stats.} Number of examples in train, dev, and test splits for our three different experimental setups. Full are the original unaltered datasets. Wikidata Answerable keeps only examples where at least one question entity and  answer entity are mappable to Wikidata and there is at least one fact between them in our set of facts.
    \label{tab:data_stats}}
\end{table}

We evaluate our model on two Open-domain Question Answering datasets: FreebaseQA \cite{jiang2019freebaseqa} and WebQuestionsSP \cite{webqsp} (See table \ref{tab:data_stats} for data statistics). Both datasets are created from Freebase. To align with our pretraining task, we convert the entity ids from Freebase to Wikidata. 

\noindent \textbf{FreebaseQA.} FreebaseQA is derived from TriviaQA and several other trivia resources (See \citet{jiang2019freebaseqa} for full details). Every answer can be resolved to at least one entity and each question contains at least one question entity $e_i$. Additionally, there exists at least one relational path in Freebase between the question entity $e_i$ and the answer $e_{\textnormal{ans}}$. The path must be either a one-hop path, or a two-hop path passing through a mediator (CVT) node, and is verified by human raters. 72\% of the question entities and 80\% of the answer entities are mappable to Wikidata, and 91.7\% of the questions are answerable by at least one answer entity that is mappable to Wikidata.

\noindent \textbf{WebQuestionsSP.} WebQuestionsSP is constructed from Freebase and contains 4737 natural language questions (3098 training and 1639 test). Questions in the dataset are linked to corresponding Freebase entities and relations. We mapped question entities and answer entities to their Wikidata ids. 87.9\% of the questions are answerable by at least one answer entity that is mappable to Wikidata. 

\noindent \textbf{Subset of questions answerable by KB triples.} Both of these datasets were constructed so that that all questions are answerable using the FreeBase KB, which was last updated in 2016.  Because our pretraining corpus is derived from larger and more recent versions of Wikipedia,  we elected to use  a KB constructed from Wikidata instead.  Use of the more recent Wikidata KB means that some questions are no longer answerable using the KB, so we also created a second reduced version of the datasets called \textit{Wikidata Answerable}.  These subsets only contains questions that are answerable by triples from our Wikidata-based KB. The model should learn to rely on the KB to answer the questions. 

\subsection{Pretraining \label{sec:results_full}}

\begin{table*}[t!]
    \small
    \centering
    \begin{tabular}{l|cc|cc}
\toprule
& \multicolumn{2}{c|}{FreebaseQA} &   \multicolumn{2}{c}{WebQuestionsSP} \\ \midrule
Data        & Full      &    WikiData                                    & Full  &     Wikidata  \\ 
        &   Dataset    &    Answerable                                    & Dataset  &     Answerable  \\ 
\midrule
FOFE        &  37.0                      &  -                                    &  67.6     & -                \\
EmQL        &  -                        &  -                                     &  \textbf{75.5}  &  74.6 \\
EaE         &  \gray{(51.4)} 53.4      &  \gray{(57.3)} 59.1                     &  46.3         &  61.4        \\
FaE (ours)  &  \gray{(61.8)} \textbf{63.3}   &  \gray{(70.1)} \textbf{73.9}      &  56.1         &  \textbf{78.5}        \\
\hline
EaE no finetune   &  \gray{(17.8)} 18.3       &   \gray{(24.2)} 24.8                                       & 12.8 & 21.4 \\
FaE (ours) no finetune    &  \gray{(20.1)} 19.7      &   \gray{(27.0)} 26.9                                       & 15.9  & 24.6 \\

\bottomrule
    \end{tabular}
    \caption{\textbf{Conventional Setting Evaluation}. Accuracy on FreebaseQA and WebQuestionsSP datasets. We pretrain our models on the full unfiltered Wikipedia text corpus. In the Full Data column, we report scores on the original unfiltered data splits for each dataset. In the WikiData Answerable column, we filter each split to only contain examples where at least one question and answer entity are mappable to WikiData and our WikiData knowledge graph contains some fact connecting them. Nearly all FreebaseQA and WebQuestionsSP entity pairs that are mappable to WikiData co-occur in the Wikipedia pretraining text. Models marked ``no finetune'' were not finetuned.}
    \label{tab:fulldata}
\end{table*}



FaE is pretrained on Wikipedia and Wikidata.
Text in Wikipedia is chunked into 128 token pieces. To compute the entity-linking loss loss$_{ent}$, we use as training data entities linked to the 1 million most frequently linked-to Wikidata entities. 
Text pieces without such entities are dropped. This results in 30.58 million text pieces from Wikipedia. As described in $\S$ \ref{sec:model_definition}, we generate $n$ training examples from a piece of text containing $n$ entity mentions, where each mention serves as the masked target for its corresponding example, and other entity mentions in the example are treated as context entities\footnote{We additionally mask context entities randomly with probability .15}. This conversion results in 85.58 million pre-training examples. The knowledge base $\mathcal{K}$ is a subset of Wikidata that contains all facts with subject and object entity pairs that co-occur at least 10 times on Wikipedia pages.\footnote{This leads to more KB triples than entity pairs, since a pair of subject and object entities can be associated with more than one relation.} This results in a KB containing 1.54 million KB triples from Wikidata (or 3.08 million if reverse triples are included). Below, this is called the \textit{full setting} of pretraining---we will also train on subsets of this example set, as described below. We pretrain the model for 500,000 steps with the batch size 2048, and we set $k=1$ in the $\textnormal{TOP}_k$ operation for fact memory access. 

\subsection{Results}

We compare FaE with three baseline models: FOFE \cite{jiang2019freebaseqa}, EmQL \cite{sun2020guessing}, and Entity-as-Expert (EaE) \cite{fevry2020entities}. FOFE is a feed-forward language model designed to encode long sequences and was the previous state-of-the-art model on the FreebaseQA dataset. EmQL was introduced as a query embedding on knowledge bases and is the previous state-of-the-art model on WebQuestionsSP. EaE has been discussed above, and our EaE models are trained using the same hyperparameters and optimization settings as FaE in order to make them as comparable as possible.

Table \ref{tab:fulldata} compares the FaE model to the baseline model. With the full pre-training and fine-tuning datasets, we outperform the baseline models on the FreebaseQA dataset by nearly 10 points. Performance on WebQuestionsSP in the Full Dataset setting is relatively lower, however this is largely explained due to the incompleteness of the KB due to mapping between Freebase and Wikidata---only 51.3\% of the questions in WebQuestionsSP are answerable using facts from our KB. In contrast, both FOFE and EmQL have complete coverage as they both use the full applicable subset of Freebase.  

However, if we instead consider only questions answerable using our dataset (the column labeled ``Wikidata Answerable'') FaE substantially outperforms EmQL. In this case, both models have complete knowledge base coverage. Additionally, in the Wikidata Answerable setting in FreebaseQA, the gap between EaE and FaE grows even larger to nearly 14 points.

Interestingly, EaE and FaE even answer many questions correctly without any fine-tuning at all (denoted ``no finetune'' in the tables. Both models answer around a quarter of the answerable questions for both datasets in this zero-shot setting with FaE having a slight advantage.




\section{Modifiable Knowledge Base}

\subsection{Filtering to Avoid Pretrain, Finetune, and Test Overlap} \label{sec:filter_overlap}

We are interested primarily in the ability of models to use external knowledge to answer questions, rather than learning to recognize paraphrases of semantically identical questions.  Unfortunately, analysis of the two datasets showed that many of the test answers also appear as answers to some training-set question: 
this is the case for 75.0\% of answers in the test data for FreebaseQA, and 57.5\% of the answers in WebQuestionsSP. This raises the possibility that some of the performance of the models can be attributed to simply memorizing specific question/answer pairs, perhaps in addition to recognizing paraphrases of the question from its pretraining data.

To resolve this issue, we discard questions in the training data that contain answers which overlap with answers to questions in the dev and test data. We end up with 9144/2308/3996 data (train/dev/test) in FreebaseQA and 1348/151/1639 data in WebQuestionsSP.
This setting is referred to as \textit{Fine-tune} column in table \ref{tab:filter_data} which shows the effects of different filterings of the data. The column denoted None has no filtering and is the same as the Full Dataset setting in table \ref{tab:fulldata}. In the column labeled \textit{Pretrain}, for every question answer entity pair in our finetuning dataset (coming from any split), we filter every example from our Wikipedia pretraining corpus where those pair of entities co-occur. Additionally, we filter every fact from our fact memory containing any of these entity pairs. In this way, the model will be unable to simple memorize paraphrases of question answer pairs that it observed in the text. Finally, the \textit{All} column combines both pretrain and fine tune filtering. We see that the models perform substantiall worse when these filterings are applied and they are forced to rely on the ability to reason across multiple examples, and in the case of FaE, the fact memory.

\begin{table*}[]
    \small
    \centering
    \begin{tabular}{l|cccc|cccc}
\toprule
& \multicolumn{4}{|c|}{FreebaseQA} & \multicolumn{4}{c}{WebQuestionsSP} \\
\midrule
Filter Type & None  &  Pretrain  & Fine-tune  &All & None  &  Pretrain  & Fine-tune  & All \\ 
\midrule
EaE               &   \gray{(51.4)} 53.4    & \gray{(42.9)} 45.2 &  \gray{(43.9)} 45.8    & \gray{(28.1)} 28.6  &  46.3   &    45.4    & 30.9      & 29.4 \\ 
FaE (ours)       & \gray{(61.8)} {63.3}     & \gray{(56.8)} 57.5  &  \gray{(53.7)} 56.5    & \gray{(47.9)} 48.0   &  56.1    & 55.4     & 40.7       & 39.2    \\
\bottomrule
    \end{tabular}
    \caption{\textbf{Effects of Different Data Filtering}. The column denoted \textit{None} has no filtering and is the same as the Full Dataset setting in table \ref{tab:fulldata}. \textit{Pretrain} removes all entity pair overlap between the eval datasets (all splits) and the pretraining text and kb. The \textit{Fine-tune} column removes all entity pair overlap between the eval train and test splits. The \textit{All} column combines both pretrain and fine tune filtering. }
    \label{tab:filter_data}
\end{table*}

\subsection{Injecting New Facts into Memory \label{sec:results_inject}}
Because our model defines facts symbolically, it can in principle reason over new facts injected into its memory, \emph{without retraining any parameters of the model}. To test how well the model is able to perform this task in practice, we look at how well the model can perform given full knowledge, filtered knowledge, and injected knowledge. The gap between the filtered knowledge setting and injected knowledge setting should demonstrate how well the model is able to incorporate newly introduced facts.


The results are shown in Table \ref{tab:inject_facts}. We always use the filtered \textit{Finetune} subset of the data (see \S \ref{sec:filter_overlap}) to avoid overlap between finetuning train and test data. In the ``Full'' column, we pretrain and finetune the FaE model with the full knowledge base and corpus. In the ``Filter'' setting, facts about the finetuning data are hidden from the model at both pretraining and finetuning time. In this case, the model should fall back to the language model to predict the answer. As shown in Table \ref{tab:inject_facts}, the performance of FaE and EaE are close. In the ``Inject Facts'' setting, Facts are hidden at pretraining time, but are injected at test time. The results show that FaE can effectively use the newly injected facts to make prediction, i.e. an absolute improvement of 9.3\%  compared to the ``Filter'' setting. EaE does not have a natural mechanism for integrating this new information\footnote{There are various heuristics one could apply for finetuning a standard language model on this type of data by applying one or a small number of gradient steps on textualized facts. We are currently exploring to what extent this is effective and what knowledge is lost during that additional learning.}.

\begin{table*}[t!]
    \small
    \centering
    \begin{tabular}{l|ccc|ccc}
\toprule
 & \multicolumn{3}{c}{FreebaseQA}                            & \multicolumn{3}{|c}{WebQuestionsSP} \\
\midrule
 & Full  &   Filter & Inject Facts             & Full  &   Filter & Inject Facts\\
 \midrule
EaE               &   \gray{(43.9)} 45.8     & \gray{(28.1)} 28.6  & \gray{(28.1)} 28.6     &   30.9    & 29.4     &      29.4  \\ 
FaE (ours)       & \gray{(53.7)} {56.5}     &\gray{(42.2)} 38.7  & \gray{(47.9)} 48.0             &  40.7       &  32.3   &      39.2   \\
\bottomrule
    \end{tabular}
    \caption{\textbf{Injecting New Facts}. In the Full setting the model is exposed to full knowledge in the pretraining data and KB. In the Filter setting, the models have access to no direct knowledge about question answer entity pairs from either the pretraining corpus or KB. In the Inject Facts setting, the pretraining corpus and training KB are still Filtered, but at inference time, new facts are injected into the models memory allowing it to recover most of the drop from the Full setting. In all cases, we remove the overlap between the finetune train and eval sets. }
    \label{tab:inject_facts}
\end{table*}

\subsection{Updating Stale Memories \label{sec:results_update}}
One of the main motivations for our model is to address the need for knowledge representations that can avoid stale data by incrementally updating as the world changes. To probe this ability, we simulate an extreme version of this scenario where all answers to QA pairs in the FreebaseQA test set are replaced with plausible alternatives. For each QA pair, we replace the original answer entity $e_{\textnormal{original}}$ with another entity from our vocabulary $e_{\textnormal{new}}$ that has 1) been used as an object in at least one of the same relation types that $e_{\textnormal{original}}$ was an object in, and 2) shares at least three Wikipedia categories with $e_{\textnormal{original}}$. We use the same pretrained models from section \ref{sec:results_full}. We fine-tune the filtered FreebaseQA train set and perform early stopping on the unmodified FreebaseQA dev set. Overall, FaE is able to utilize the modified KB to make the correct prediction for ~30\% of questions. 

While this is an encouraging result, the decrease in performance compared to the unmodified evaluation set (nearly twice as many incorrect predictions) shows that the mixing between conextual representations and knowledge requires further research. In section \ref{sec:results_inject} FaE was able to easily adapt to using newly injected facts because they were consistent with the pretraining corpus. These were facts that did not appear in the model's pretraining data but they also did not contradict that data. In the case of updating stale memories, we are instead giving the model new information that in some cases (such as in this experiment) explicitly contradict the knowledge stored in its latent parameters, and this inconsistency makes the mixing much more difficult. Addressing this issue as well as the even more difficult problem of deleting knowledge is a main focus of ongoing and future research.



\section{Conclusion \label{sec:conclusion}}
In this paper, we presented a method for interfacing a neural language model with an interpretable symbolically bound memory. We used that interface to change the output of the language model by modifying only the non-parametric memories and without any additional training. We demonstrated the effectiveness of this method by performing comparably or better than a high performing language model on factoid question answering while integrating new facts unseen in pretraining data. We even showed that we can modify facts, such that they contradict the initial pre training text, and our model is still largely able to answer these questions correctly.

\bibliographystyle{acl_natbib}

\end{document}